\documentclass[letterpaper, 10 pt, conference]{ieeeconf}  
\usepackage{cite}
\usepackage[utf8]{inputenc} 
\usepackage[T1]{fontenc}    
\usepackage{hyperref}       
\usepackage{url}            
\usepackage{booktabs}       
\usepackage{amsfonts}       
\usepackage{nicefrac}       
\usepackage{microtype}      
\usepackage{xcolor}         
\usepackage{booktabs}       
\usepackage{amsfonts}       
\usepackage{nicefrac}       
\usepackage{microtype}      
\usepackage{xcolor}         
\usepackage{graphicx}
\usepackage{multicol}
\usepackage{amsmath}
\usepackage{multirow}

\IEEEoverridecommandlockouts                              

\usepackage{booktabs}                                   
\usepackage{multirow}                                   
\usepackage{makecell}                                   
\usepackage{tablefootnote}                              
\usepackage[symbol]{footmisc}                           
\usepackage{amsmath,amssymb}                            
\usepackage{xcolor}                                     
\let\labelindent\relax              
\usepackage{enumitem}                                   
\usepackage{subcaption}                                 
\usepackage{stfloats}                                   
\usepackage[misc]{ifsym}     
\usepackage{hyperref}   

\newcommand{\eat}[1]{}                                  

\title{\LARGE \bf
Let Me Show You:
Learning by Retrieving \\ from Egocentric Video for Robotic Manipulation
}

\author{Yichen Zhu$^{1}$, Feifei Feng$^{1}$
\thanks{$^{1}$Midea Group, AI Research Center, China. 
        {\tt\small \{zhuyc25, feifei.feng\}@midea.com}}
}
\begin{document}

\maketitle

\thispagestyle{empty}
\pagestyle{empty}

\begin{abstract}
Robots operating in complex and uncertain environments face considerable challenges. Advanced robotic systems often rely on extensive datasets to learn manipulation tasks. In contrast, when humans are faced with unfamiliar tasks, such as assembling a chair, a common approach is to learn by watching video demonstrations. In this paper, we propose a novel method for learning robot policies by Retrieving-from-Video (RfV), using analogies from human demonstrations to address manipulation tasks. Our system constructs a video bank comprising recordings of humans performing diverse daily tasks. To enrich the knowledge from these videos, we extract mid-level information, such as object affordance masks and hand motion trajectories, which serve as additional inputs to enhance the robot model's learning and generalization capabilities. We further feature a dual-component system: a video retriever that taps into an external video bank to fetch task-relevant video based on task specification, and a policy generator that integrates this retrieved knowledge into the learning cycle. This approach enables robots to craft adaptive responses to various scenarios and generalize to tasks beyond those in the training data. Through rigorous testing in multiple simulated and real-world settings, our system demonstrates a marked improvement in performance over conventional robotic systems, showcasing a significant breakthrough in the field of robotics. 
\end{abstract}

\section{Introduction}

\begin{figure}[t]
    \centering
    \includegraphics[width=0.5\textwidth]{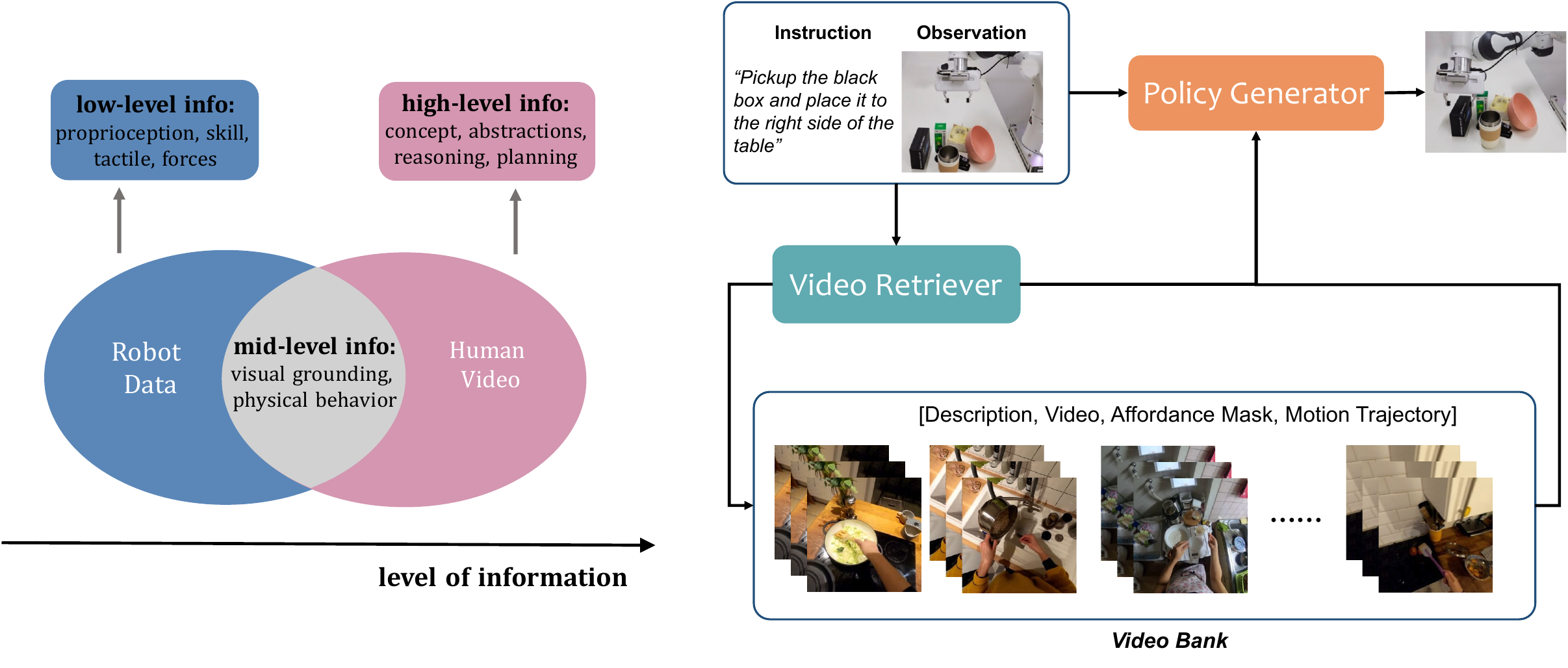}
    \caption{\textbf{Left:} The level of information that we gain from robot data and video \textbf{Right:} The overview of our retrieving-from-video framework.}\label{fig:init}
\end{figure}

The advancement of foundation models in areas like natural language processing and computer vision has sparked interest in the robotics community to create embodied agents capable of comprehending human instructions and responding aptly to their environment. Despite this enthusiasm, crafting agents that seamlessly interact with the physical world remains a formidable task. Deep neural networks typically contain a large number of neurons, enabling the implicit storage of knowledge extracted from vast amounts of data. However, recent studies in robotics~\cite{zhu2024language, zhu2024retrieval, wen2024object, jiang2024ragraph, wen2025tinyvla, wen2025diffusionvla, wen2025dexvla, wang2024visualrobot, li2025pointvla, zhu2025scaling, zhou2023make, huang2022label, li2024mmro, wu2025discrete, zhu2025objectvla, li2025coa,meng2023logsummary,zhu2022teach,zhu2024retrieval1,biglog,zhu2021unilog,wen2024object1,zhou2023make1,zhou2025vision,huang2022label2,li2024improving,wu2024discrete1,10907650li2024visual,huang2022labelarxiv, wen2024object2,zhu2021unilog1,zhou2025chatvla2,zhu2024retrieval2}
demonstrate that scalability in terms of both training data and model size falls short~\cite{zitkovich2023rt2, bousmalis2023robocat} when compared to foundation models in other domains, such as Large Language Models. This insight has inspired the creation of robot models designed to learn efficiently with limited data and model sizes. To augment their capabilities, it's increasingly important for these robots to access external repositories of mid-level knowledge, such as visual dynamics, physical behavior, and language grounding, as demonstrated in Figure~\ref{fig:init}. This knowledge thereby expands their capacity to understand and interact with the world.

The ability to tap into an external repository of behavioral memory mirrors the human learning process. Imagine a scenario where a child with no prior experience in handiwork, receives an IKEA table set and wishes to assemble it independently. A common solution is to watch a video demonstrating the assembly process and then mimic the steps shown. This ability is crucial for performing tasks that require task-specific knowledge and learning from a new environment. Consequently, the question naturally arises: How can we harness the wealth of videos demonstrating human actions to enhance the precision of robots in manipulation tasks? This inquiry not only explores the potential of robotic learning but also seeks to bridge the gap between human learning processes and robotic optimization procedures.

In this paper, we present Retrieving-from-Video (RfV), a method that enables robots to learn manipulation tasks by observing human demonstrations. Plain human videos with language descriptions may contain high-level information, such as abstraction and reasoning of the scene, which may not be directly beneficial for robotic manipulation. To address this, we extract mid-level information, including object affordance and motion trajectory, that can be helpful for robots to learn low-level controls. Figure~\ref{fig:init} (left) provides a summary that defines the level of information used in our approach.

To realize our objectives for learning by retrieving, we introduce two modules, a video retriever and a policy generator. The video retriever module retrieves task-relevant videos from the video bank based on language instructions from a human user. This enables us to obtain videos that closely match the current tasks. The policy generation module effectively integrates mid-level information to facilitate both the training and testing of policy networks. During model training, the retrieved video, along with its mid-level information, serves as an additional data point to enhance the robot's learning process. At test time, the videos retrieved from the bank act as in-context samples that help the model adapt to dynamic environments. The overview of our framework is presented in Figure~\ref{fig:init} (right).

The efficacy of our proposed Retrieving-from-Video (RfV) framework is demonstrated through extensive evaluation across multiple simulation benchmarks and real-world experiments. This approach showcases the versatility and practical ability of our framework.

In summary, our contributions are as follows:

\begin{itemize}
    \item We introduce a novel RfV method that leverages knowledge from human videos to enhance policy learning for robots. Our pipeline extracts mid-level information from videos, boosting the robot model's performance.
    \item Our framework features a video retriever that retrieves task-relevant videos based on human language instructions and a policy generator that integrates this additional knowledge to improve robotic manipulation.
    \item We validate our methodology through extensive evaluations in both real-world settings and multiple simulated environments. The results strongly affirm the effectiveness and practicality of our approach.
\end{itemize}

\section{Methodology}

\begin{figure*}[t]
    \centering
    \includegraphics[width=0.7\textwidth]{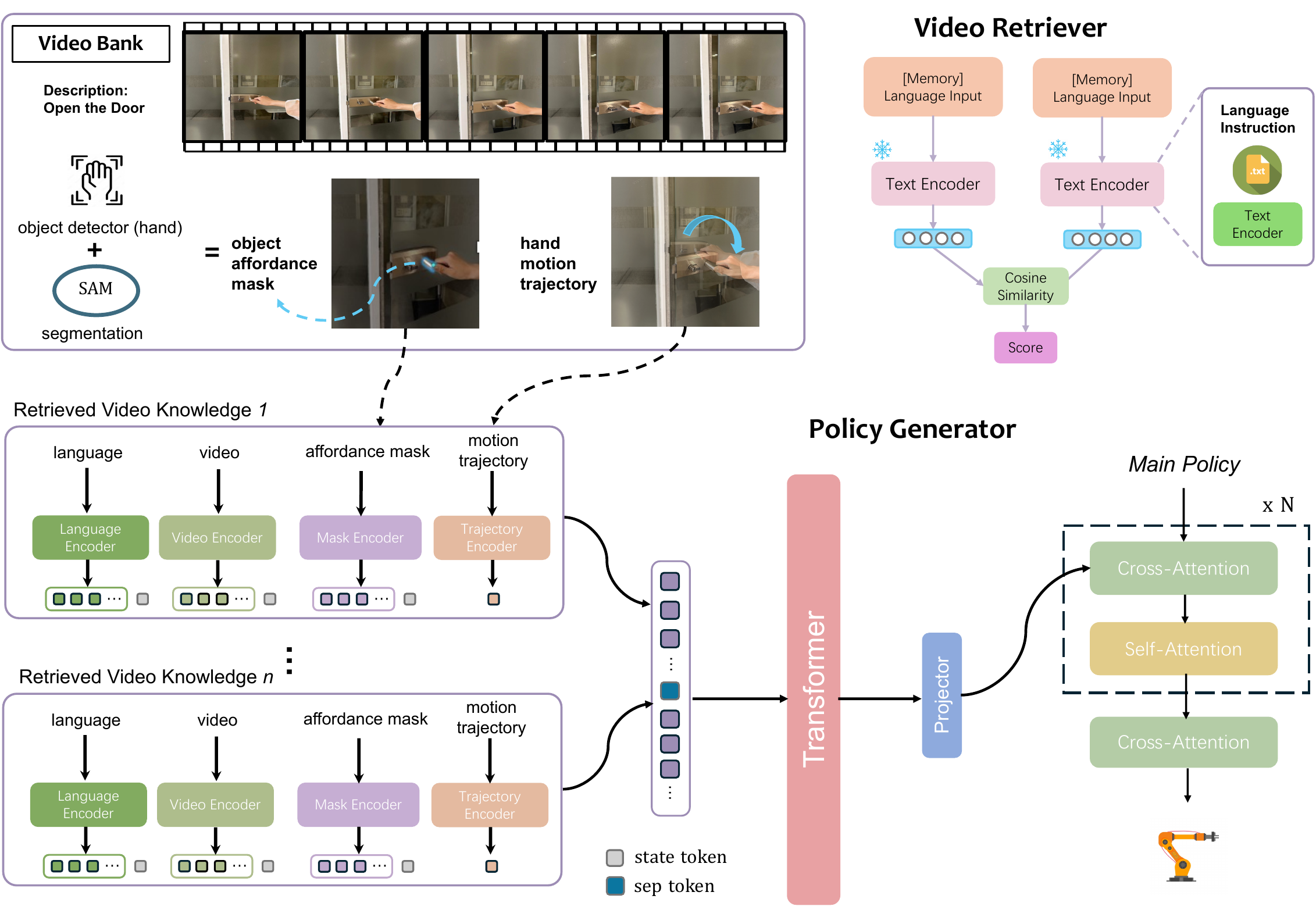}
    \caption{The framework of our RfV consists of three main components: the video bank (top left), the video retriever (top right), and the policy generator (bottom). The video retriever retrieves relevant videos based on language instructions, while the policy generator processes the retrieved videos and their mid-level information to facilitate the training and evaluation of the robot model.}\label{fig:rfv_framework}
\end{figure*}

We introduce Retrieving-from-Video (RfV) for the robotics framework in this section. Our method builds a video bank and then retrieves videos that humans conducting the same task and fuses the information into policy networks. An illustration of our framework is presented in Figure~\ref{fig:rfv_framework}. 

\subsection{Preliminaries}

\textbf{Notations}. In our setting, we assume access to a video dataset $D_{video}$. A video clip is denoted as $v = (s_{0}, s_{1}, \cdots, s_{T})$. Here, we denote a video clip as $v= (s_{0}, s_{1}, \cdots, s_{T} )$, where $v \in D_{video}$ is the full clip and each $s$ is an observation in the form of an image. We use $i_{\text{video}}$ to represent the language-based narrations of the video and $i_{\text{robot}}$ as the language instruction for the robot. In our framework, we extract the object affordance represented as masks $\alpha$ and hand motion trajectory $\tau$. 

The framework consists of a video retriever $R$ and a policy generator module $G$. The retrieval module $R$ takes an input sequence $i_{\text{robot}}$ and searches the $i_{\text{video}}$ from an external video bank. If $i_{\text{robot}}$ and $i_{\text{video}}$ achieve high similarity, it returns a list of video information 
$m = \{i_{\text{robot}}, v, \alpha, \tau\}$. The policy generator $G$ then takes the input sequence $x \in D_{robot}$ and multiple retrieved video information $M = \{m_{1}, m_{2}, \cdots, m_{n}\}$ and returns the action $a$, where $a$ represent continuous actions that control the robots.

\subsection{Constructing Mid-Level Information from Video}
One key challenge is building the video banks. While plain human videos can be useful for robot learning, they often contain redundant information that may misguide the training process and introduce additional computational costs. Moreover, these videos lack mid-level information, such as visual dynamics, which can be beneficial for robot training. To address this, we focus on extracting mid-level information that is helpful for robotic manipulation from human videos. Specifically, we extract the object affordance map and hand motion trajectory. Our data annotation pipeline, described below, enables us to extract this mid-level knowledge from human videos. This process is executed offline and does not impact model training and inference efficiency.

Consider a video $v$ consisting of $T$ frames. We have a dual objective: to identify the location of contact, and to determine the subsequent movement of the hand. The contact point, also known as the affordance map, represents the areas of objects that humans can manipulate. The estimation of hand movement, referred to as the hand motion trajectory, instructs the robot on how to maneuver post-grasping. This mid-level information is crucial for enabling the robot to interact with real-world objects and helps it generalize to new environments using only human demonstrations.

We initiate our process by pinpointing the keyframe in the video where the human hand contacts the object. To construct the object affordance map, we utilize the open-vocabulary object detector GroundingDino~\cite{liu2023groundingdino}, which first localizes the position of the hand. Following this, we employ GPT-4V to ascertain the name of the object currently held by the hand. Finally, we deploy Segment Anything (SAM)~\cite{kirillov2023segmentanything}, using a text prompt and the pixels around the hand, to precisely define the affordance mask.

The pixel-space position of the hand forms the post-grasping trajectory, denoted as $\tau$. To delineate contact points, we compute the centroid of the bounding box across all frames to construct the hand motion trajectory. This trajectory can be visualized by plotting these points or vectors on each frame, or by overlaying a continuous path on the video. However, in the real world, the camera often moves over time, and the bounding box might be inaccurate, leading to a jittery raw trajectory. To address this, we apply a smoothing algorithm, specifically spline interpolation, to achieve a cleaner and more realistic trajectory.

In our study, we utilize the Ego4D datasets~\cite{grauman2022ego4d} as our primary video repository. Given that most robotic manipulation tasks occur indoors, we exclude videos captured outdoors. To classify each video as indoor or outdoor, we analyze the first frame using GPT-4V. This step effectively filters out videos that do not align with the typical environments found in standard robotic manipulation benchmarks. Nevertheless, we acknowledge that videos filmed outdoors could be valuable for training robotic models intended for outdoor activities.

\subsection{Video Retriever}
A video retriever $R$ takes in a task specification query $q$, which is typically a language instruction, and obtains a video $M$ from the video bank $M$. Then, a relevance score can be obtained through our model. We follow prior retrieval works~\cite{karpukhin2020dense}, in which the retriever $r$ is a bi-encoder architecture,
\begin{equation}
    r(q, m) = E_{Q}(q)^{T}E_{M}(m)
\end{equation}
We employ two key encoders: $E_{Q}$, responsible for encoding queries, and $E_{M}$, which encodes memory to produce dense vectors representing the query and memory policies, respectively. In our scenarios, the task specification is provided as a language instruction. Thus, we utilize the text encoder from CLIP~\cite{clip}, which is trained on image-text pairs, to extract feature representations and compare the similarity between the query and memory. The CLIP model is both compact and efficient, facilitating the rapid comparison of feature similarities for retrieval purposes. For the retrieval process, we perform Maximum Inner Product Search within the memory space, generating a ranked list of candidates based on their relevance scores. From this list, we select the top $k$ videos for further analysis and processing. The video is retrieved in both the training and test stages to ensure that the learning process is consistent. 

Furthermore, it is common practice to use multiple cameras to provide the robot with more comprehensive visual information for manipulation tasks. For each view, we independently retrieved a video clip from the video bank. We found that this straightforward approach was sufficient for enabling the model to learn in-context knowledge from the retrieved videos. One possible improvement could be to use a video generation method to create additional viewpoints based on a single view, which we leave for future work.

\begin{figure}[t]
    \centering
    \includegraphics[width=0.5\textwidth]{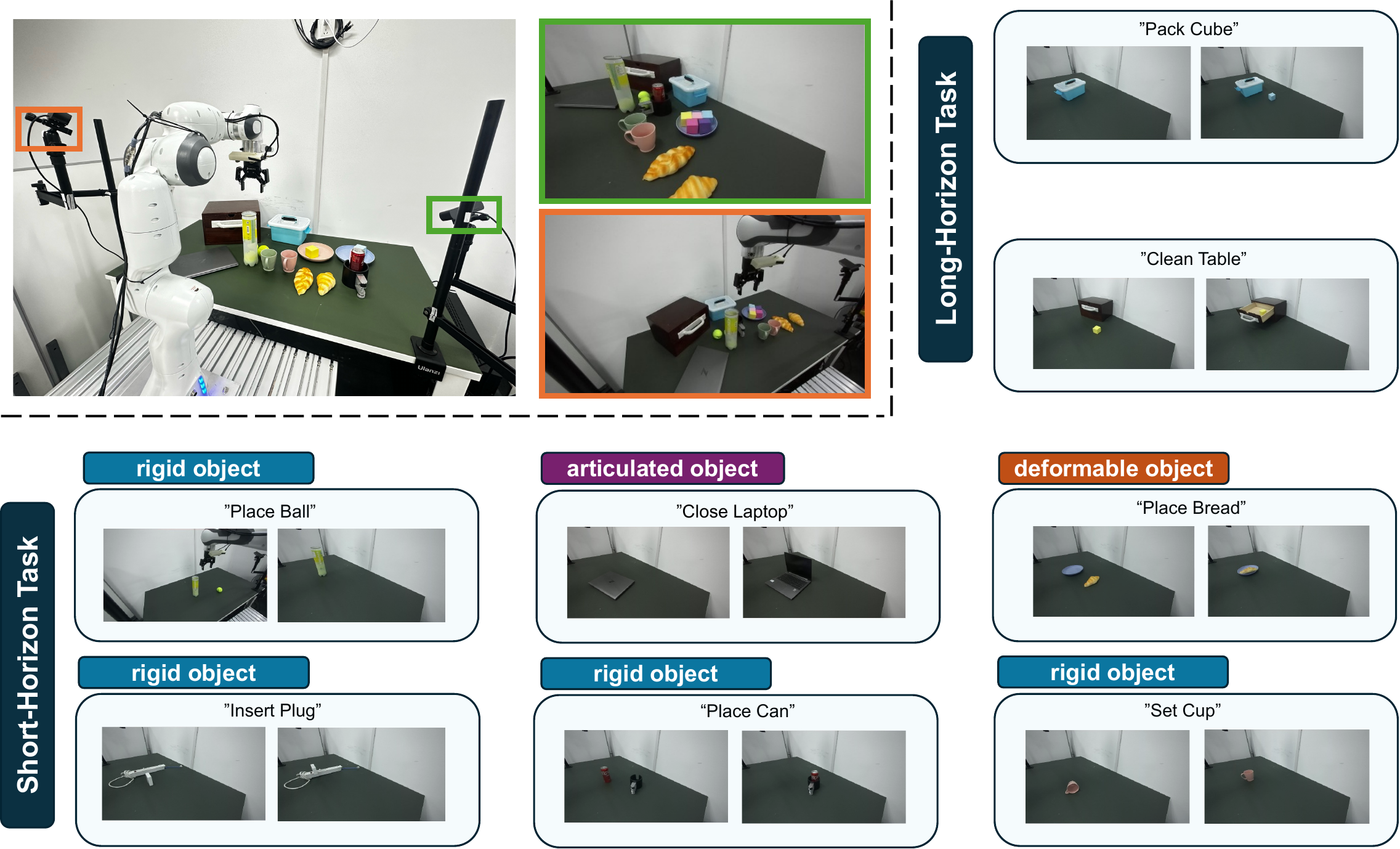}
    \caption{The setup of our Franka real robot and the example of tasks in our real-world experiments.}\label{fig:workspace}
\end{figure}
\subsection{Policy Generator}
The policy generator is designed to effectively utilize the valuable information in the retrieved policy to facilitate the training of the policy for the current input. First of all, we reuse the feature representation of the text encoder from the video retriever. Then, for human video, we utilized an image-based model to get frame-wise features in the form of tokens. To be specific, we utilize pre-trained ViT-Base as our visual feature extractor. To further improve computational efficiency, we conduct memory consolidation by merging the most similar tokens in the adjacent frames following ToMe~\cite{tome}. By reducing 90\% of the visual tokens, our transformer model is as fast as using a single image. For the affordance mask and hand motion trajectory, we utilize a mask encoder and a trajectory encoder, both composed of multi-layer perceptrons (MLPs), to generate corresponding tokens. These tokens are concatenated with text and video tokens, facilitating an effective combination of video data in the training of robot data. We introduce a learnable state token between the text and video tokens to distinctly separate the data from two different states. For simplicity, we call these tokens as video feature tokens $M$.

Given a list of retrieved video feature tokens $M = (m_{1}, ..., m_{K})$, we concatenate these tokens according to their relevance scores and use absolute position embeddings to preserve the order of tokenized representations. A "sep" token is utilized to distinguish between tokens from different policies. Once concatenated, the tokens are processed using the Transformer architecture. We enhance the integration of the retrieved features $M$ into the policy network by employing cross-attention mechanisms. We use projection layers for the retrieved videos to act as query and key, with the robot data feature representation serving as the value. This setup ensures that the valuable representations from retrieved videos are effectively utilized in the main network, enhancing policy learning for the current input. For the main policy network, we follow the design choice of Action Chunking Transformer (ACT)~\cite{act} to train the policy conditioned on current observation via behavior cloning. 
\section{Experiments}

\subsection{Simulation Experiments}

\begin{table}[t]
    \centering
    \caption{Experimental results of Bi-Dexhands~\cite{chen2022towards} and Metaworld~\cite{yu2020metaworld}, a simulation benchmark. The numbers in parentheses indicate the number of tasks for the simulation benchmark.}
    \label{tbl:simulation_experment}
    \resizebox{0.5\textwidth}{!}{\begin{tabular}{c|cccc}
        \toprule
        \multirow{2}{*}{Method} & \multicolumn{4}{c}{Metaworld}\\
         & Easy (28) & Medium (11) & Hard (6) & Very Hard (6) \\
        \midrule
        VINN~\cite{parisi2022unsurprising} & 20.6 & 5.2 & 2.7 & 0.0\\
        BeT~\cite{shafiullah2022behavior} &24.5 & 9.1 & 0.9 & 0.0\\
        ACT~\cite{act} &47.6 & 15.4 & 4.8 & 8.4\\
        Diffusion Policy~\cite{chi2023diffusion} & 82.1 & 35.4 & 15.6 & 12.3 \\
        \midrule
        \textbf{Ours} & \textbf{93.6} & \textbf{54.5} & \textbf{21.8} & \textbf{15.7} \\
        \bottomrule
    \end{tabular}}
\end{table}
In this section, we evaluate our approach using Metaworld~\cite{yu2020metaworld}, a widely-used simulation benchmark.

\noindent
\textbf{Evaluation}: For the simulation benchmark, we evaluate on Metaworld~\cite{yu2020metaworld} Medium level and Hard level, following the settings in MWM~\cite{seo2023masked}. All experiments were trained with 30 demonstrations and evaluated with 3 seeds, and for each seed, the success rate was averaged over five different iterations.

\noindent
\textbf{Experimental Results.} In our studies, we carried out a comparative analysis using the Retrieving-from-Video approach. We evaluated our method against several leading imitation learning approaches, including BeT~\cite{shafiullah2022behavior}, VINN~\cite{parisi2022unsurprising}, ACT~\cite{act} and Diffusion Policy~\cite{chi2023diffusion}. Our policy network was trained through a few-shot learning method, using datasets that included ten demonstrations. The results, illustrated in Table~\ref{tbl:simulation_experment} for the Meta-World benchmarks, clearly show that our method outperforms the baseline methods in terms of effectiveness. Notably, our method significantly outshines the others. For instance, on Metaworld medium-level benchmarks, our proposed Retrieving-from-Video outperforms the Diffusion Policy by 19.1\% and the Action Chunking Transformer (ACT) by 39.1\%. Moreover, in challenging tasks where baseline methods exhibit low success rates, such as Hard and Very Hard tasks,, our method leads by a large margin. These findings confirm the superior performance of our approach, particularly in leveraging video demonstrations for retrieval in few-shot and challenging scenarios.

\subsection{Real-Robot Experiments}
We show that our proposed Retrieving-from-Video can learn to perform precise manipulation tasks, obtain good performance with very few training data, and obtain generalizability in terms of appearance, spatial, and many others. 

\textbf{Experimental Setup.} Our real-world experiment was conducted using a Franka Emika robot across eight distinct tasks. We utilized two ZED cameras to capture real-world visual observations. One camera was positioned on the left side of the robot, while the other was placed on the right side, ensuring comprehensive visual coverage. We visualized our experimental setup and all tasks in Figure~\ref{fig:workspace}. We now briefly describe our tasks:

\begin{table*}[t]
\renewcommand{\arraystretch}{1.3}
\caption{Experiments on real robot. Our method consistently outperforms Baseline in all environments. All metrics are reported in percentage $(\%)$ with the best ones bolded. The symbol * denotes pretraining on 970K OpenX~\cite{padalkar2023openx} robot data.}
\centering
\resizebox{\textwidth}{!}{\begin{tabular}{l|ccccccccc}
\toprule
Model & PlaceBread & PlaceBall & PlaceCan & CloseLaptop & SetCup & InsertPlug & CleanTable & PackCube  & Avg. \\
\midrule
R3M~\cite{nair2022r3m} & 30 & 0 & 0  & 20  &  0 & 5 & 10 & 0 &  8.1 \\
VIMA~\cite{jiang2022vima}  & 70 & 5  & 0  & 45 & 60 & 30 & 30 & 15 & 30.6\\
RT-1~\cite{brohan2022rt1} &  80 & 5  & 5  & 80 & 45 & 50 & 30 & 35 & 41.3 \\
Octo*~\cite{team2024octo} & 50 & 25 & 30 & 50 & 35 & 40 & 25 & 45 & 37.5\\
OpenVLA*~\cite{kim2024openvla} & \textbf{90} & \textbf{65} & 50 & 75 & 25 & 30 & 40 & 60 & 54.4\\
Diffusion Policy~\cite{chi2023diffusion} & 80 & 20 & 10 & 90 & 35 & 25 & 50 & 40 & 43.8 \\
ACT~\cite{act} & \textbf{90} & 55 & 40 & \textbf{100} & 55 & 65 & 45 & 60 & 63.8\\ 

\midrule
RfV & \textbf{90} & \textbf{65} & \textbf{60} &  \textbf{100}  &  \textbf{75} & \textbf{70} & \textbf{60} & \textbf{65} & \textbf{73.1} \\
\bottomrule
\end{tabular}}
\label{table:real_world_exp}
\end{table*}

\begin{table*}[t]
\renewcommand{\arraystretch}{1.3}
\caption{Ablation study on the real robot. Our experiments demonstrate that the mid-level information is crucial to the success of our method.}
\centering
\resizebox{\textwidth}{!}{\begin{tabular}{l|ccccccccc}
\toprule
Model & PlaceBread & PlaceBall & PlaceCan & CloseLaptop & SetCup & InsertPlug & CleanTable & PackCube  & Avg. \\
\midrule
RfV & \textbf{90} & \textbf{65} & \textbf{60} &  \textbf{100}  &  \textbf{75} & \textbf{70} & \textbf{60} & \textbf{65} & \textbf{73.1}\\
- hand motion trajectory & 85 & 45 & 50 &  85  & 60 & 65 & 40 & 40 & 58.8 \\
- object affordance & 80 & 35 & 30 &  80  &  30 & 50 & 35 & 25 & 45.6 \\

\bottomrule
\end{tabular}}
\label{table:ablation_mid}
\end{table*}

\begin{table}[t]
\renewcommand{\arraystretch}{1.3}
\caption{Ablation study on the real robot. Our experiments demonstrate that the mid-level information is crucial to the success of our method. The number indicates retrieved video for one view.}
\centering
\resizebox{0.45\textwidth}{!}{\begin{tabular}{l|cccc}
\toprule
Number of Retrieved Videos & 1 & 3 & 5 & 7 \\
\midrule
Avg. Success Rate & 46.4 & 73.1 & 70.0 & 72.8\\

\bottomrule
\end{tabular}}
\label{table:ablation_number_video}
\end{table}

\textbf{Experimental Results} 
We perform multiple studies to delve into various questions related to our model's performance and capabilities. All models are trained with 50 demonstrations for each task, and all models are trained with the same number of training iterations. We compare our method with multiple state-of-the-art methods, including R3M~\cite{nair2022r3m}, VIMA~\cite{jiang2022vima}, RT-1~\cite{brohan2022rt1}, Octo~\cite{team2024octo}, OpenVLA~\cite{kim2024openvla}, Diffusion Policy~\cite{chi2023diffusion}, and ACT~\cite{act}. Notice that OpenVLA and Octo are pre-trained on 970K OpenX robot data. 

\noindent
\textit{1. How effective are Retrieving-from-Video?} In In Table~\ref{table:real_world_exp}, we present experimental results from eight real-world tasks. We compare our Retrieving-from-Video (RfV) framework with prominent foundational robot models such as R3M~\cite{nair2022r3m}, VIMA~\cite{jiang2022vima}, and RT-1~\cite{brohan2022rt1}. Our findings indicate that RfV consistently outperforms these state-of-the-art methods across all tasks. Specifically, in tasks involving the manipulation of rigid objects (PlaceBall and PlaceCan), our method achieves an average success rate of 40\% and 30\%, respectively, surpassing the capabilities of other models, which struggle to complete these tasks. Similarly, for long-horizon tasks like CleanTable and PackCube, RfV also demonstrates a higher success rate compared to alternative approaches. These results align with our earlier simulations, further validating the effectiveness of our method.

\noindent
\textit{2. How important is mid-Level information?} We further explore the impact of integrating mid-level information, including hand motion trajectory and object affordance, on task performance. As shown in Table~\ref{table:ablation_mid}, removing either motion trajectory or object affordance significantly impacts the overall success rate across all tasks. In particular, the absence of motion trajectory data results in a substantial decrease in success rates for tasks requiring intricate movements, such as CloseLaptop and various long-horizon tasks. Likewise, the removal of object affordance data leads to lower success rates in tasks that demand precise manipulation, such as PlaceBall and PlaceCan. These findings highlight the critical role of mid-level information in ensuring successful task completion.

\begin{figure*}[t]
    \centering
    \includegraphics[width=\textwidth]{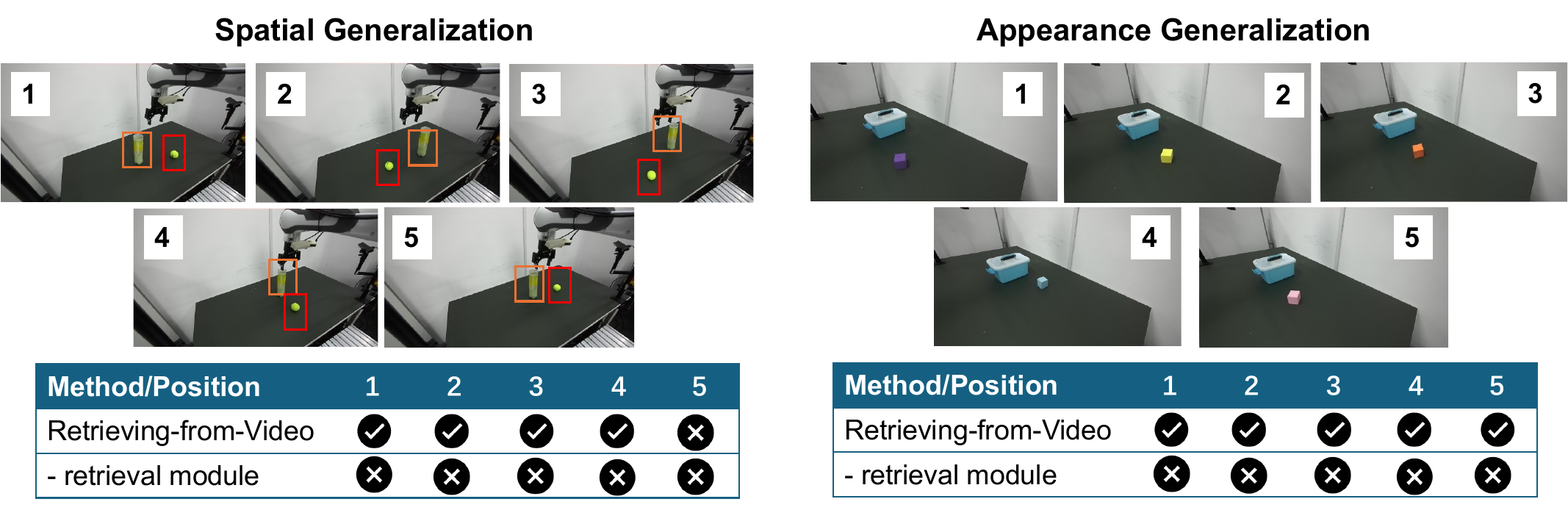}
    \caption{\textbf{Left:} The spatial generalization experiments setup. We randomly placed the tennis ball (highlighted by red bounding box) and tennis ball box (highlighted by orange bounding box). \textbf{Right:} The appearance generalization. We change the color of the cube, which is not presented in the training data.}\label{fig:rfv_general}
    
\end{figure*}

\textit{3. How does the number of retrieved videos affect model performance?} The key idea of our paper is video retrieval. This section explores the impact of the number of retrieved videos on model performance. In Table~\ref{table:ablation_number_video}, we evaluate four different settings. Our results indicate that retrieving three videos yields the best performance. Beyond this point, increasing the number of retrieved videos leads to a slight decline in performance. Thus, we conclude that retrieving three videos is optimal for our method.

\textit{4. Does Retrieving-from-Video improve robot generalizability?}
Besides the effectiveness in handling all tasks, RfV shows strong generalization abilities in the real world. We categorize the generalization abilities of RfV into 3 aspects and detail each aspect as follows. 
\\
\\
\noindent
\textbf{Spatial generalization.} In Table~\ref{table:real_world_exp}, we present quantitative results demonstrating the spatial generalization of our model, specifically when objects are randomly positioned as described in our dataset. We assess whether our model can generalize to object positions that were not encountered during training. As shown in Table~\ref{fig:rfv_general} (left), our RfV model successfully generalizes to new object positions in 4 out of 5 trials. In contrast, removing the retrieval module results in the model's complete inability to generalize to any of the test positions. We attribute this to the auxiliary mid-level information provided by our retrieval module, which significantly enhances the model’s capability to grasp objects and execute movement based on language instructions.
\\
\\
\noindent
\textbf{Distractor generalization.} Conventional robot models often lack robustness against distractors. To assess the distractor generalization capability of our method, we tested whether it could successfully complete manipulation tasks in the presence of distractors that were not included in the training data. Specifically, we introduced five different distractors during the PlaceBall tasks, including plastic bottles, glass cups, toy bears, headphones, and keyboards. When we removed the retrieval module, the success rate dramatically fell from 80\% (4 out of 5 successes) to 0\%. These results underscore the importance of incorporating our proposed retrieval module to enhance the robot's policy generalization in environments containing unseen distractors.
\\
\\
\noindent
\textbf{Appearance generalization.} We evaluate the appearance generalization of our approach by providing specific language instructions regarding the color of the cube, such as "place the yellow/red/blue cube." Conventional robot learning methods typically fail when the color of the objects changes, as demonstrated in Table~\ref{fig:rfv_general} (right). The model without a retrieval module fails because these models can only recognize objects that were present in the training data. In contrast, our retrieval-from-video method can generalize to novel colors, succed 5 out of 5. This is because our retrieval method enables the model to understand the mapping between the color description and the actual appearance of the object, and then apply this understanding to learning. More importantly, the primary objective of this work is to demonstrate that our method can effectively generalize without the aid of any data augmentation, thereby underscoring the potential of the RAG approach in real robot learning.

\section{Conclusion}
In this study, we introduce Retrieving-from-Video (RfV), a novel framework that capitalizes on the plethora of human video data to enhance robotic manipulation performance. RfV demonstrates exceptional proficiency across a broad spectrum of robotic tasks in both simulated and real-world settings. The fundamental advantage of RfV lies in its ability to incorporate strategically curated mid-level information from human videos, thereby enriching the expressiveness and efficacy of policy learning. In real-world applications, it achieves high precision in complex manipulations involving both articulated and rigid objects and adeptly manages long-horizon tasks. Overall, our methodology presents an innovative and practical approach to leveraging human video as an external source for robot learning. 

\bibliographystyle{IEEEtran}
\bibliography{reference}

\end{document}